\documentclass[a4paper]{article}
\usepackage{sty/artefact}
\usepackage{sty/bib_nat}
\usepackage{bibentry}
\usepackage{mfirstuc} 
\blendcolors*{!20}
\usepackage{xspace}
\graphicspath{{fig/}}

\author{Glen Hopkins*, Kristjan Kalm \\
\normalsize{Artefact UK, London}\\
\normalsize{glen.hopkins@artefact.com, kristjan.kalm@artefact.com}
}
\title{Classifying complex documents: comparing bespoke solutions to large language models}
\begin{document}
\setcounter{page}{1}
\maketitle

\newcommand{\numstates}{12\xspace}
\newcommand{\numcounties}{267\xspace}
\newcommand{\numdocs}{29,307\xspace}
\newcommand{\numtest}{600\xspace}

Here we search for the best automated classification approach for a set of complex legal documents. Our classification task is not trivial: our aim is to classify ca 30,000 public courthouse records from \numstates states and \numcounties counties at two different levels using nine sub-categories. Specifically, we investigated whether a fine-tuned large language model (LLM) can achieve the accuracy of a bespoke custom-trained model, and what is the amount of fine-tuning necessary. We first give a brief summary of the main results and then detail the data, tasks, and models.

\subsection*{Summary}

We contrasted three approaches: (1) a bespoke, custom-trained classification model; (2) large language model (GPT-3.5); and (3) a fine-tuned large language model (GPT-3.5).
We found that a bespoke classification model outperformed LLM-based approaches (Fig 1). However, fine-tuning a LLM drastically improved its performance, whilst not reaching the performance levels of the bespoke model (Fig 1, right panel).

\begin{figure}[!ht]
    \centering
    \includegraphics[width=0.98\linewidth]{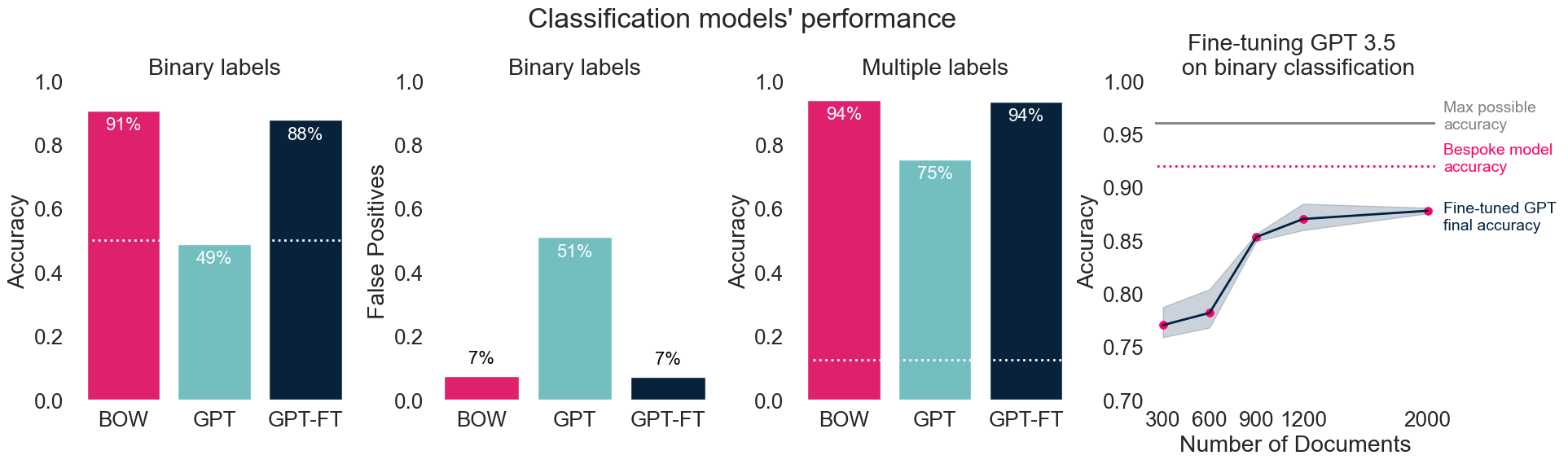}
    \caption{Accuracy breakdown for three classification models. (1) Binary classification accuracy (details in \textit{Tasks}). BOW: bespoke bag-of-words classification model; GPT: GPT-3.5; GPT-FT: fine-tuned GPT-3.5. The white dashed line represents chance-level performance. (2) Proportion of false positives in the binary task. (3) Multi-label classification accuracy. (4) Fine-tuned GPT performance on binary classification as a function of the number of documents used for fine-tuning. The grey line represents the maximum performance achievable (0.96) due to noisy data. The red dashed line represents the performance of the bespoke, BOW classification model. The x-axis shows the number of documents used to fine-tune GPT, with the shaded area around the line representing standard deviation across measurements.} 
    \label{fig:enter-label}
\end{figure}

Our bespoke approach combined a bag-of-words model of n-grams with a convolutional neural net, and was trained on \numdocs example documents (see \textit{Models} for details). 
Our results suggest that while such custom-trained models achieve the highest accuracy, fine-tuning LLMs leads to comparable, but lower accuracies, representing a good value for effort. However, LLM's performance gains from fine-tuning plateau after initial improvements (Fig 1, right panel), and the accuracy does not reach the levels of bespoke approaches.

\subsection*{Background}

Document classification is usually the first step in any document processing workflow. It's common for big organisations in both corporate and public sectors to process hundreds of thousands to millions of complex documents within a calendar year. However, classification of complex legal, medical, or supply-chain documents is still done semi-manually in many organisations, since the costs of misclassification are high, as sending a document to a wrong workflow is costly. Here we searched for the best automated classification approach for a set of complex legal documents. We contrasted two approaches: a bespoke, custom trained classification model, versus a large language model fine-tuned with significantly fewer data. 

\subsubsection*{Classification challenge}
Our classification task is not trivial: we are looking at a set of ca 30,000 public courthouse records from \numstates states and \numcounties counties. At the first level, we are interested only in the legal documents relevant to us, and discard the rest. This is a binary classification task independent of the legal type of the document: for example, we only want to keep lease agreements that pertain to oil and gas production and discard the rest. The information for the decision is not usually in the title (e.g. ”Agreement of Lease”) or inferred from document properties (date, parties, terms, etc.) but contextually present somewhere in the text. After the binary classification, we want to further break down the documents based on their legal properties: e.g. extensions, releases, corrections, addendums, memorandums, etc, comprising further nine sub-categories. In sum, the classification task has two levels, binary and multi-label classification, and requires processing the text of the document.

\subsubsection*{Data}
\begin{itemize}
    \item \numdocs documents of courthouse records across the USA (\numstates states and \numcounties counties).
    \item Binary classification (2 labels): label the text as \colorbox{atf_red}{"Oil and Gas Document"} or \colorbox{RoyalBlue}{"Other"}. The \colorbox{atf_red}{"Oil and Gas Document"} label was used to represent any legal document that deals with the production, distribution, or storage of oil and gas related commodities or assets. The \colorbox{RoyalBlue}{"Other"} label was assigned to the remainder of the documents.
    \item Multi-class classification (9 labels): all \colorbox{atf_red}{"Oil and Gas Documents"} were further classified into nine exclusive categories: \colorbox{atf_red}{"Affidavit of Non-Production"}, \\ \colorbox{atf_red}{"Affidavit of Production"}, \colorbox{atf_red}{"Assignment of Oil and Gas Lease"}, \\ \colorbox{atf_red}{"Correction"}, \colorbox{atf_red}{"Extension"}, \colorbox{atf_red}{"Memorandum of Lease"}, \colorbox{atf_red}{"Oil and Gas Lease"}, \\ \colorbox{atf_red}{"Release"}, and \colorbox{atf_red}{"Top Lease"}.
    \item Dataset labelling and quality: true label values were assigned to the training set documents manually by a team of domain experts. Subsequent quality control by random sampling revealed that ca 4-6\% of the labels were incorrect, resulting in the maximum possible achievable accuracy of 96\% for any model. This reflects well the noisy nature of most real-life complex datasets.
    \item All models were tested with the same set of \numtest documents. Training set varied across the models: the bespoke classification model was trained with \numdocs labelled documents.
\end{itemize}

\subsection*{Models}

Over the last few years, the standard implementation of complex classification tasks has required labelling a sizeable training set (usually in the tens of thousands), encoding the documents’ texts numerically, and feeding the encodings to a customisable model, such as a convolutional neural net, to predict the category of each document, based on the similarities between the encodings. In the field of natural language processing (NLP) research, several robust and highly accurate classification algorithms have been developed, based on the distributions of words and phrases in the text. However, such a classic NLP approach can be costly to implement: labelling thousands of documents precisely requires a lot of manual labour by domain experts. Furthermore, achieving a highly accurate and a robust classifying solution requires model training and optimisation work by data scientists.

\subsubsection*{Large language models: fine-tuning for specialist tasks}
Over the last year, large language models (LLM, such as GPT, Llama, Mistral, and Bard amongst others) have become quickly affordable and robust enough to be considered viable candidates for complex text processing tasks. OpenAI’s GPT is probably the most popular and widely-used, gaining high adoption rates in professional services, and spanning a whole sub-industry of LLM-powered tools. However, such models are trained on vast quantities of domain-general data, and thus can lack the precision and resolution required to perform specialist tasks. LLMs also tend to “hallucinate” in tasks with high uncertainty. To address these issues, LLM developers have set up frameworks to fine-tune their models for specific tasks: a task-specific context is first set up, and then a labelled dataset used to train the LLM for that particular task only. This approach has seen fine-tuned LLMs to achieve significantly higher accuracy on complex domain-specific tasks than before, such as medical prediction, legal reasoning and, contract understanding \cite{li2023, guha2023legalbench, hendrycks2021cuad}. 

Although fine-tuning still requires manual data labelling, the size of the training data is usually significantly smaller by an order of a magnitude and the fine-tuning process itself delegated to the LLM provider via an existing framework (such as an API), which in turn, significantly reduces the workload of data scientists involved. In sum, the cost of setting up a highly accurate and domain-specific classification model with fine-tuning LLMs can be potentially also an order of a magnitude lower than with a bespoke NLP model, whilst still achieving similar performance levels.

\subsubsection*{Bespoke NLP approach -- an ensemble neural net model}

We used an approach which is an ensemble of two models — a bag-of-words (BOW) model based on n-gram encoding, and a convolutional neural network with an attention mechanism. The ensemble model first tokenises the text using n-gram encoding, then creates feature embeddings based on the occurrence frequency of the n-grams, which is finally fed into a single hidden-layer neural net with a soft-max output layer. 

\begin{figure}[ht]
    \centering
    \includegraphics[width=0.9\linewidth]{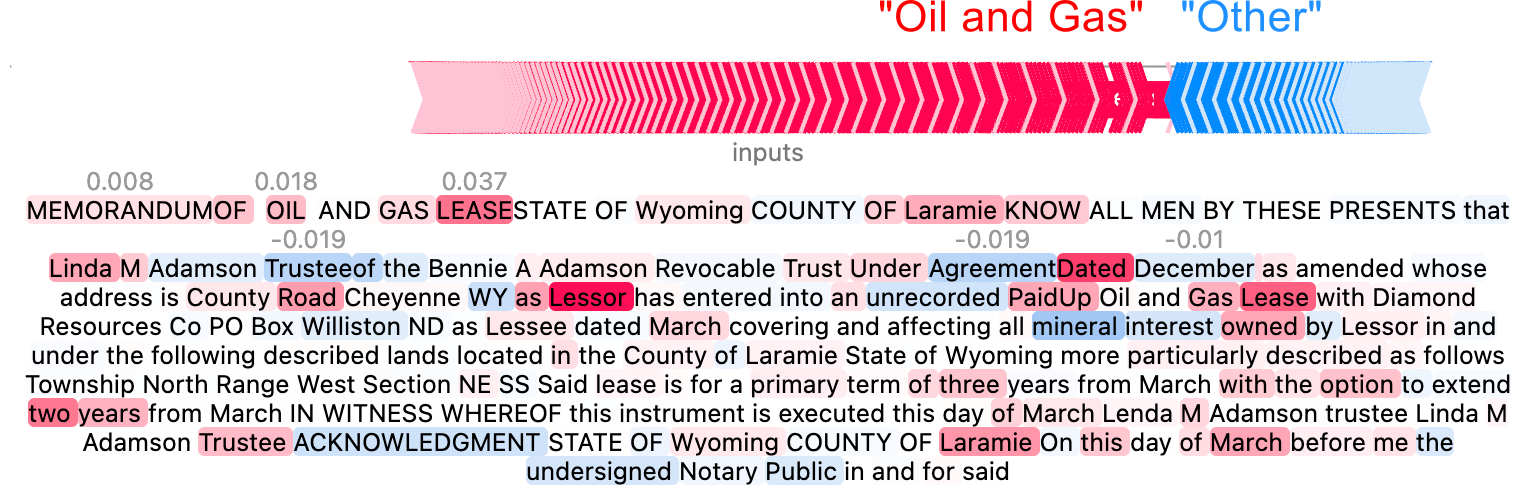}
    \caption{Neural net ensemble model: the "bag-of-words" mechanism.}
Shapley values of text tokens: the importance of each token to the classification decision is overlaid on the original text that corresponds to that token.
    \label{fig:shapley_explainer}
\end{figure}

The n-gram bag-of-words (BOW) model represents the whole text document as a "bag of n-grams", rather than sequentially organised text. In other words, this model is only concerned in whether certain phrases (of n-length) occur in a document or not, and completely discards any information about the structure, context, and the appearance order of the phrases. 

Since to-be-classified documents could be up to hundreds of pages long, we had to limit the length of the input text to the model. For this purpose, we conducted a hyperparameter search during training to determine what is the optimal context window (in terms of maximising expected accuracy) for both binary and multi-label classification. In practical terms, this meant determining the optimal length of the text fed to the classifying model per document. The optimal text length was 800 tokens for binary, and 1500 tokens for the multi-class task.

\subsubsection*{LLM: GPT-3.5}
\begin{itemize}
    \item GPT (Generative Pre-trained Transformer) 3.5 is a large language model (LLM) released by OpenAI in 2022. We use the gpt-3.5-turbo-1106 version \cite{gpt}.
    \item We used GPT-3.5 API to predict both binary and multi-class labels for a set of \numtest documents. We presented the to-be-classified text with a task to choose one exclusive value from either a binary or a multi-label taxonomy.
\end{itemize}

\subsubsection*{Fine-tuned LLM: Chat-GPT-3.5}
\begin{itemize}
    \item GPT-3.5 API provides an interface to fine-tune the model by exposing it to specific data with an aim of increasing performance on a particular task. 
    \item During fine-tuning we provided examples of document text, the document class and a prompt for the task itself.
    \item In the binary task, we used the prompt: \textit{"Classify documents as either 'Oil and Gas Document' or 'Other' based on text entered."}
    \item For the multi-class we used the prompt: \textit{"Classify documents as one of the following: 'Affidavits of Non-Production', 'Affidavits of Production', 'Assignment of Oil and Gas Lease', 'Correction Documents', 'Extension', 'Memorandum of Lease', 'Oil and Gas Lease', 'Releases', 'Top Lease'."}
    \item We used different numbers of documents to fine-tune GPT-3.5: 200, 600, 900, 1200, 2000. This was done to measure the improvement in classification performance as a function of fine-tuning set size.
\end{itemize}

\subsection*{Conclusions}

We have shown that fine-tuning an LLM with significantly fewer data (2,000 vs. 30,000 documents) leads to lower, but comparable performance levels to a bespoke NLP model on a complex classification task. Organisations should definitely consider a fine-tuned LLM as a first prototype solution for a document classification problem, as it sets a baseline of value-for-money. However, there are several important reasons why a bespoke NLP solution might be preferable.

\textit{Lack of control}: a business-critical process will be dependent on a third-party service, with related uncertainty over pricing, availability, and security. Furthermore, third parties change their models over time: for example, there are significant differences in response behaviour between older and newer versions of GPT-3.5 \cite{chen2023chatgpts}. Although organisations can set up their own LLMs to be fine-tuned and utilised in a very similar way, the cost of maintaining and training an in-house LLM is significantly higher than any associated with a bespoke NLP model. 

\textit{Data sharing}: if the data cannot be shared with a third party, LLM fine-tuning becomes difficult. This is a deal-breaker for many organisations in medical and legal industries. As before, running an in-house LLM will break the cost-effectiveness of the approach.

\textit{LLMs require a validation and post-processing layer}: this is necessary to control for possible “hallucinations” and add business logic to catch outliers and validate outcomes. For example, one of our fine-tuned models predicted a document class of \textit{Subordination of Oil and Gas Lease}, which is not a document type included in our multi-class classification task. In short, a follow-up processing pipeline is required to make the solution production-ready, and this is usually easier to implement for a transparent bespoke solution.

Lastly, \textit{not all NLP tasks are good fits with LLMs}: while the classification task is a naturally good fit with a generative text models (e.g. classification can be considered as a specific case of generating a summary) that is not true for highly-constrained NLP tasks such as extracting rigidly defined information from unstructured texts. For example, GPT 3.5 cannot yet compete with bespoke NLP models for named entity recognition (NER) or structured relation extraction \cite{zhou2023universalner}. As a general strategy, a purely LLM-based architecture is not yet feasible for all NLP tasks, especially for complex processing chains. However, we expect this to change significantly over the next year as LLMs become involved in the specific stages of larger NLP tasks. For example, a dominant NLP framework Spacy has recently integrated LLM-support as a customisable component \cite{spacyllm} of their complex processing pipelines.

\bibliography{doc}

\end{document}